\documentclass[lettersize,journal]{IEEEtran}
\usepackage{amsmath,amsfonts}
\usepackage{algorithmic}
\usepackage{algorithm}
\usepackage{array}
\usepackage[caption=false,font=normalsize,labelfont=sf,textfont=sf]{subfig}
\usepackage{textcomp}
\usepackage{stfloats}
\usepackage{url}
\usepackage{verbatim}
\usepackage{graphicx}
\usepackage{cite}
\begin{document}

\title{Inverse kinematics learning of a continuum manipulator using limited real time  data}

\author{Alok Ranjan Sahoo, \IEEEmembership{Member, IEEE}, Pavan Chakraborty, \IEEEmembership{Member, IEEE}
\thanks{This paper was produced by the IEEE Publication Technology Group. They are in Piscataway, NJ.}
\thanks{Manuscript received April 19, 2021; revised August 16, 2021.}}

\markboth{Journal of \LaTeX\ Class Files,~Vol.~14, No.~8, August~2021}%
{Shell \MakeLowercase{\textit{et al.}}: A Sample Article Using IEEEtran.cls for IEEE Journals}

\IEEEpubid{0000--0000/00\$00.00~\copyright~2021 IEEE}

\maketitle

\begin{abstract}
Data driven control of a continuum manipulator requires a lot of data for training but generating sufficient amount of real time data is not cost efficient. Random actuation of the manipulator can also be unsafe sometimes. Meta learning has been used successfully to adapt to a new environment.  Hence, this paper tries to solve the above mentioned problem using meta learning. We consider two cases for that. First, this paper proposes a method to use simulation data for training the model using MAML(Model-Agnostic Meta-Learning). Then, it adapts to the real world using gradient steps.  Secondly,if the simulation model is not available or difficult to formulate, then we propose a CGAN(Conditional
Generative adversial network)-MAML based method for it. The model is trained using a small amount of real time data and augmented data for different loading conditions. Then, adaptation is done in the real environment. It has been found out from the experiments that the relative positioning error for both the cases are below 3\%. The proposed models are experimentally verified on a real continuum manipulator. 
\end{abstract}

\IEEEPARstart{L}{ack} of enough degrees of freedom and safety issues of the conventional manipulators lead researchers to the continuum manipulators. Snakes \cite{snakeinspiredrobot,virgala2021snake,pettersen2017snake}, tentacles \cite{Jorgensen2022,fras2018fluidical,martinez2013robotic}, elephant trunks \cite{hannan2003kinematics,wu2021fem,BHA:2014,Yang:2006} were primary sources of inspiration for them. Mimicking their flexible designs is not an easy task.   

Hence, researchers tried uniform backbone designs first. It was having simpler mathematical model but independent actuation of each section was difficult with it. Hence, tapered designs were introduced. Model based controllers were used for it. However, the mathematical formulations of it were complex. The properties of the manipulator must be known for it. Hence, data driven approaches were explored to reduce the difficulty.

In terms of model accuracy, it has been demonstrated that data-driven approaches perform better than model-based approaches. RNN based control \cite{thuruthel2018stable, thuruthel2017learning}, reinforcement learning\cite{you2017model}, deep reinforcement learning \cite{satheeshbabu2019open} are being used for the continuum manipulator. However, the requirement of high amount of data and time creates problem in using them in real world. These models are also not so effective to adapt to slight changes in the environment. To make our model truly intelligent, our model should be able to adapt to the new environment using a very small amount of changed data. Meta learning has shown significant success to generalize among similar types of tasks\cite{MAMLoriginalpaper}. Hence, we will explore its capability to generalize with a very small amount of real time data points. 

It has been observed that meta learning based models perform well if it is trained with the data points which are similar to the real world scenario \cite{sahoo2022study,zhao2020sim,simtoreal}. If we train it from scratch, then we will still need a considerable amount of real data points. Data points are generally generated using random actuation of the actuators. This can be unsafe and unpredictable some times. Generating real time data can also be costly. To address this problem generally the model is first trained with simulation data. Then,  transfer learning \cite{zhuang2020comprehensive,tanaka2021trans}, knowledge distillation \cite{rusu2015policy}, imitation learning, meta learning are used for the its adaptation in real world. However, meta learning based sim-to-real adaptation has not been explored for continuum manipulators yet. Hence, we intend to explore meta learning for it in this paper. We will use MAML for it.

We segregated this problem in to two parts. First, if the simulation model is available, then we can use it for initial training. Then, we can adapt it using a meta learning(learning to learn) approach. Second, if the model is not available, then we can use a CGAN (Generative Adversarial Networks) model to create a data set from some small amount of real time data. Hence, we tried to find the effectiveness of the above two approaches for our highly non-linear continuum manipulator.

\subsubsection{A brief review of MAML}

Meta learning attempts to uncover generalisation after being trained with many sets of similar tasks, taking its cue from humans who tend to generalise after a few similar encounters. It is known as "learning to learn" since it tries to generalise across the questions. After training, it leverages this expertise (meta knowledge) to learn any task.

Meta learning can also be thought of as a bi-level optimization method. It alludes to a situation involving hierarchical optimization in which one optimization acts as a constraint on another optimization. The basic task such as classification or regression is done by the inner layer. The outer objective is taken care of by the meta learner. Task generalization performance or learning speed can be the outer objective function for the outer layer\cite{Meta_learning_in_neural_networks}.

MAML(Model Agnostic Meta Learning ) was proposed by finn et al.\cite{MAMLoriginalpaper}. It was proposed for quick adaptation of a new task. It can be applied to any model trained with gradient descent. It can be applied to classification, regression, reinforcement learning etc. As we have already seen it becoming successful to adapt to a new condition in our previous work, we believe it has the potential to adapt to a real environment from simulation environment. 

Let $\phi$ be the parameters of a model  $M_\phi$ and $D(\tau)$ be the distribution over the tasks( $\tau$). Let us consider that the model takes a single gradient step for the adaptation of the new task $\tau _i$. Then the update equation can be represented as 
\begin{equation}
   \phi'_i = \phi - \alpha \nabla_\phi L_{\tau_i}(M_\phi). 
\end{equation}
Where, $\alpha$ = step size and L= loss function.
The optimization is done for the model parameters based on the models performance with respect to different tasks sampled from D($\tau$). 
\begin{equation}
   argmin_\phi\sum_{\tau_i\sim D(\tau)} L_{\tau i}(M_{\phi'_i})=\sum_{\tau_i\sim D(\tau)} L_{\tau i}(M_{\phi-\alpha \nabla_\phi L_{\tau i}(M_\phi)})
\end{equation}
After that, model parameters are updated and the meta optimization is carried out using stochastic gradient descent.

\begin{equation}
   \phi \leftarrow \phi - \beta \nabla_\phi \sum_{\tau_i\sim D(\tau)} L_{\tau i}(M_{\phi'_i})
\end{equation}

Here, $\beta$ is the meta step size. We will use Mean-squared error(MSE) as our loss function. Our L will be defined as 

\begin{equation}
   L_{\tau i}(M_\phi)= \sum_{x^{(j)},y^{(j)}\sim \tau_i} ||M_\phi(x_{(j)})-y_{(j)}||_2^2
\end{equation}

where $x^{(j)}$ and $ y^{(j)}$ are an input/output pair sampled from task
$\tau_i$.


Continuum robot kinematics modelling requires two steps. First, a robot specific mapping between actuator space and configuration space is required. Then, the mapping between configuration space and task space is done. Given a target point, task space to actuator space mapping has to be done via configuration space. Here, we will try to find the mapping between task space to actuator space directly in the simulation environment and then, we will adapt it to the real world.

\subsection{Problem formulation}

We had used real data points for the initial training in our previous work\cite{sahoo2022study}. However, it is not always possible to generate sufficient amount of real data points. Hence, we need to find a technique to reduce the requirement of real data. MAML has shown great success to adapt to new conditions and environments. Hence, we propose a MAML based technique to address the above stated problem. We will do the initial training with a simulation based model. Then, we will use small amount of real data points for adaptation of the real environment. Then, we propose a CGAN(Conditional Generative adversial network) model for the data generation for the cases where developing the model of a manipulator is difficult.

To the best of the author’s knowledge, it is the first work to use meta learning for the sim-to-real adaptation. Contributions of this paper are
1. A meta learning based sim-to-real adaptation technique for a continuum manipulator where the simulation model is available.
2. A CGAN(conditional generative adversarial network) model for generating
data of different loading conditions.
3. A hybrid meta learning and generative adversarial network approach for training and adaptation of external loading was proposed.

\begin{algorithm}
\caption{Adaptation to the real environment}
\begin{algorithmic} 
\STATE  Step 1: Generate samples for learning in simulation environment using rand function (($p_{next}$,$a_{curr}$,$p_{curr}$) $\rightarrow$ $a_{next}$)
\STATE  Step 2: Train the model with simulation data using MAML 
\WHILE{not done}
       \STATE Generate samples for learning in real environment using rand function(($p_{next}$,$a_{curr}$,$p_{curr}$) $\rightarrow$ $a_{next}$)
       \STATE Update the model using the new K data samples
\ENDWHILE

\label{Algorithm2}
\end{algorithmic}
\end{algorithm}
 As explained in our previous work \cite{sahoo2022study}, a position level controller can be formulated using the following equation
\begin{equation}
  J(a_i)a_{i+1} \approx  p_{i+1}-\psi(a_i)+J(a_i)a_i
\end{equation}

The $\psi$ function in this case gives a mapping from the actuator space to the task space. The task space is $ p \in R^n$  and the actuator space is $ a \in R^m$.  $a_{i+1}$ and $a_i$ represent the present and next actuator positions respectively. Current and next position of the end effector are represented by $p_i$ and $p_{i+1}$. As feed-back to a controller has shown higher success then open loop, we will learn a mapping between $ (p_{i+1},a_i,p_i) \rightarrow a_{i+1}$. Similar to our previous work, our model will explore the whole task space. Whole range of actuators will be used for data collection. 

\section{Architecture of the neural network}

There are three hidden layers of size 128 in our neural network. There are 11 nodes in the input layer. Target tip position, present tip position, present actuator position, and external load are the inputs. The output layer is having 4 nodes. The output is the target actuator positions. ReLU was used as the activation function. The optimizer used for it was Adam. Both
 inner loop step size  and meta step size are kept at 0.01. The loss function has defined as:

\begin{equation}
   L(M)= \sum_{x^{(j)},y^{(j)}\sim \tau_i} ||(M(x_{(j)})-y_{(j)}||_2^2
\end{equation}
Here, $x^{(j)}$ is the tensor of  $ (p_{i+1},a_i,p_i)$,load and $y^{(j)}$ is the $a_{i+1}$(target tip positions). The loss function is the MSE between the predicted actuator locations and the actual actuator positions needed to reach a specific target.
\section{Simulation environment}

Here, we have used 3d mechanics based model by Camarillo et al.\cite{camarillo:2009} for the simulation. It was used because this is capable to accommodate variable length backbone. As our prototype is a spring based continuum manipulator, the backbone length varies with respect to the actuation. In our previous work, we had shown the procedure to adapt the model for our prototype. We have used the parameters obtained from the characterization of our manipulator in our previous work. 
Here, to generate the data points, we have used the forward kinematics of the above mentioned model. The external loading has been incorporated using the tension in the string as:
\begin{align*}
T_{external loading}= T_{zero loading}+ external load/4    
\end{align*}

\subsection{Simulation data collection}
We collected 10,000 different input/output pairs for external loading of 0 to 1kg(with a resolution of 0.1 kg). Rand function was used for random actuation of the actuators. Initially, we maintained a restriction that difference between the actuator position of a particular actuator in consecutive actuation should be less than 10\% of the maximum actuator length( 0.25 m). We changed the actuator values to corresponding encoder values to match the input types of the real robot. After 10\% data collection, we removed this restriction. The data set was segregated into training, validation set and test set in the ratio of 70:15:15.   

\section{Experimental setup}

We have used our spring based continuum manipulator for our experimentation\cite{sahoo2021development,sahoo2022study}. The manipulator is 0.77 long. It is having 4 sections. However, only section 2 and 4 are active sections here. The spring back-boned manipulator has a pvc skin. our previous work can be referred for the complete description of the experimental set up and the manipulator\cite{sahoo2021development,sahoo2022study} . We will be using the distal section for our experimentation. Table \ref{table1} shows the parameters used for the training.

\begin{table}
\caption{Hyper parameters used}
\label{table1}
\setlength{\tabcolsep}{3pt}
\begin{tabular}{|p{100pt}|p{100pt}|}
\hline
Step size & 0.01
\\
\hline
Meta step size & 0.01
\\
\hline
k & 10
 \\
\hline
Meta batch size & 20
\\
\hline
Number of hidden layers & 3
 \\
\hline
Hidden layer size & 128
\\
\hline
Epochs & 200
\\
\hline
\end{tabular}
\label{traingperformance}
\end{table}

\section{Collecting data from the prototype  }
We discretized the actuator space with respect to the encoder values to collect the data points from the prototype. Here, we gave 101 random actuator values for each external loading in the range of [0,0.5] with a difference of 0.05 kg. The actuator values and corresponding tip point values were recorded for the trajectories. 20 values in regular interval between the starting input-output pairs were treated as the data points for training(excluding starting point). Hence, we have 2000 data points for each external loading condition. In the following experiments, we will see the number of gradients required for a particular external loading to adapt to that external load in the real world environment. Here, we will use an external load of 0.2 kg on the prototype.

\section{Experiments and results}
\subsection{Testing with the single section model}

To test the capability of adaptation, we first trained the model with a single section continuum manipulator. The simulation model used the parameters of the distal section of the prototype. The length of the manipulator was taken to be 0.77 m. The model was a single section uniform cross section and uniform stiffness continuum manipulator. 10,000 data points were collected for the training as described in the previous section. We took random 30 points in the real world to find the average error. For adaptation purpose, we will do gradient update after 50 samples of input-output pair of the real environment. To know the actual achievable accuracy, we tested a model trained with the real data points. The accuracy achieved with the real data point was 0.0210 $\pm$ 0.0119 m. We will use this as the default accuracy.

\begin{table}[h]
\caption{Results of the model trained with real data}
\label{table}
\setlength{\tabcolsep}{3pt}
\begin{tabular}{|p{50pt}|p{50pt}|p{50pt}|p{50pt}|}
\hline
No. of data points &
Average error(in m) & standard deviation(in m) 
 \\
\hline
0& 0.0210 & 0.0111 
\\
\hline
\end{tabular}
\label{simulationresultcombined5050}
\end{table}

\begin{table}[h]
\caption{Results of the model trained with uniform continuum manipulator data(single section)}
\label{table}
\setlength{\tabcolsep}{3pt}
\begin{tabular}{|p{50pt}|p{50pt}|p{50pt}|p{50pt}|}
\hline
No. of gradient steps &
Average error(in m) & standard deviation(in m) 
 \\
\hline
0& 0.0839 & 0.0254 
\\
\hline
1 & 0.0625 & 0.0233 
\\
\hline
2 & 0.0543 & 0.0192 
\\
\hline
3 & 0.0522 & 0.0213 
\\
\hline
4 & 0.0495 & 0.0189 
\\
\hline
5& 0.0416 & 0.0173 
\\
\hline
6 & 0.0358 & 0.0155 
\\
\hline
7 & 0.0359 & 0.0154 
\\
\hline
\end{tabular}
\label{simulationresultsinglesection}
\end{table}

Here, from the table \ref{simulationresultsinglesection}, we can observe that model trained with the single section data can achieve a positioning error up to 4.64\%. We had seen this type of initial error in our last work. There, we had put an external load above the allowed external load (0.5 kg)(table \ref{Realresultsfromprotype}). The model could reduce the error up to  6.8\% in that case. However, in this case it could reduce the error up to 4.64\%. Hence, we tried to reduce the error further by using a double section model.

\subsection{Testing with the double section model}

Here, the length of each section was 0.38m. The distal section inherited the parameters of the distal section of the prototype. Similarly, the proximal section of the simulation model inherited the parameters of the second section. We again followed the same procedure as followed for the single section case. 

\begin{table}[h]
\caption{Results of the model trained with two section continuum manipulator data}
\label{table}
\setlength{\tabcolsep}{3pt}
\begin{tabular}{|p{50pt}|p{50pt}|p{50pt}|p{50pt}|}
\hline
No. of gradient steps &
Average error(in m) & standard deviation(in m) 
 \\
\hline
0& 0.0774 & 0.0244 
\\
\hline
1& 0.0406 & 0.0248
\\
\hline
2& 0.0344  &  0.0181 
\\
\hline
3& 0.0289 &  0.0139
\\
\hline
4& 0.0256 &  0.0128
\\
\hline
5& 0.0209 &  0.0115
\\
\hline

\end{tabular}
\label{simulationresultdoublesection}
\end{table}
Table \ref{simulationresultdoublesection} describes the results obtained for different number of gradient steps. We can observe from the table that the initial error of the model is slightly less than the previous model. However, it could reduce the relative mean error to the range of 2.75\% with 5 gradient steps. Hence, we can say that we can use the controller trained with the double section model in real world .  

\subsection{Random point target with different loading condition}

Now, let us see the capability of the model to adapt to the external load in the real world directly after the adaptation step in the last experiment. We will give 50 random points as the targets for each external loading. Then,we will find the average error for it. 

\begin{table}[h]
\caption{Results  for the different loading conditions}
\label{table}
\setlength{\tabcolsep}{3pt}
\begin{tabular}{|p{50pt}|p{50pt}|p{50pt}|p{50pt}|}
\hline
External loading (in kg) & Average error (in m) & standard deviation (in m) 
 \\
\hline
0 & 0.0211 & 0.0139 
\\
\hline
0.1 & 0.0214 & 0.0124 
\\
\hline
0.2 & 0.0207 & 0.0113
\\
\hline
0.3 & 0.0221 & 0.0133
\\
\hline
0.4 & 0.0216 & 0.0141
\\
\hline
0.5 & 0.0219 & 0.0138
\\
\hline
\end{tabular}
\label{randommodel-1}
\end{table}



\begin{figure}[]
\centerline{\includegraphics[width=\columnwidth]{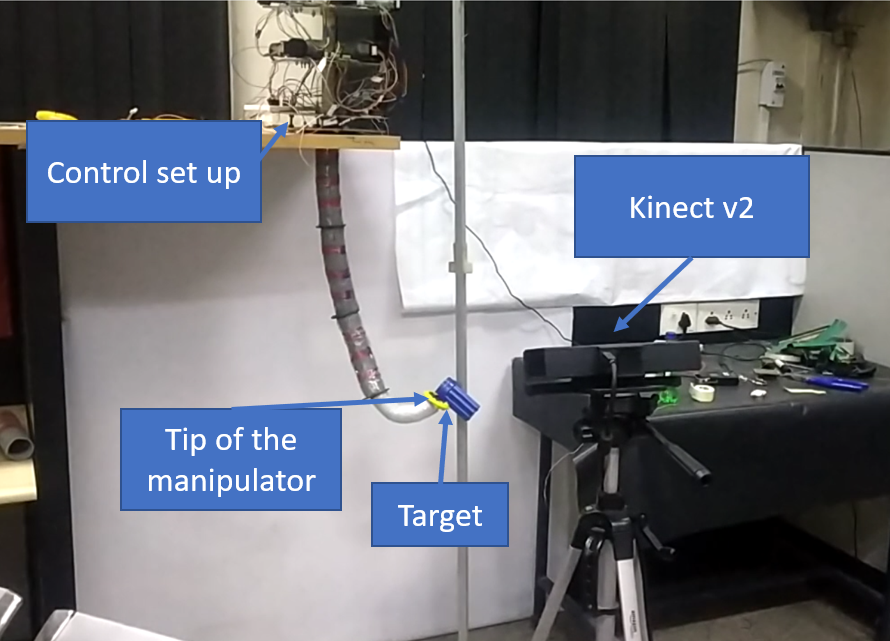}}
\caption{Target reaching by the manipulator in real environment}
\label{targetreaching}
\end{figure}
From the table \ref{randommodel-1}  we find that the model performs equally well for all the loading conditions. This gave best result for 0.2 kg case as the model had adapted to the same external loading. However, the average error of the model did not not have much difference between different loading conditions. It shows that if a model is trained with data points of different loading conditions, then it will perform much better than the approach we took in our last work. Figure \ref{targetreaching}
shows the capability of the manipulator to reach the target.

\subsection{Trajectory following experiment with known loading}

To see the capability of the model trained with the simulation data, we will consider two different trajectories. The first one will be a straight line and second one will be a semi circle. The length of the straight line is 0.60 m. We took 18 control points in between starting and ending positions with an interval of 0.03 m. The radius of the semicircle was chosen to be 0.3m. Control points at an interval of 10 degrees were given for it. Here, we have used an external loading of 0.2 kg.

\begin{figure}[htp]
	\centering
	\subfloat[]{\label{straightline0shot:1}\includegraphics[width=\columnwidth]{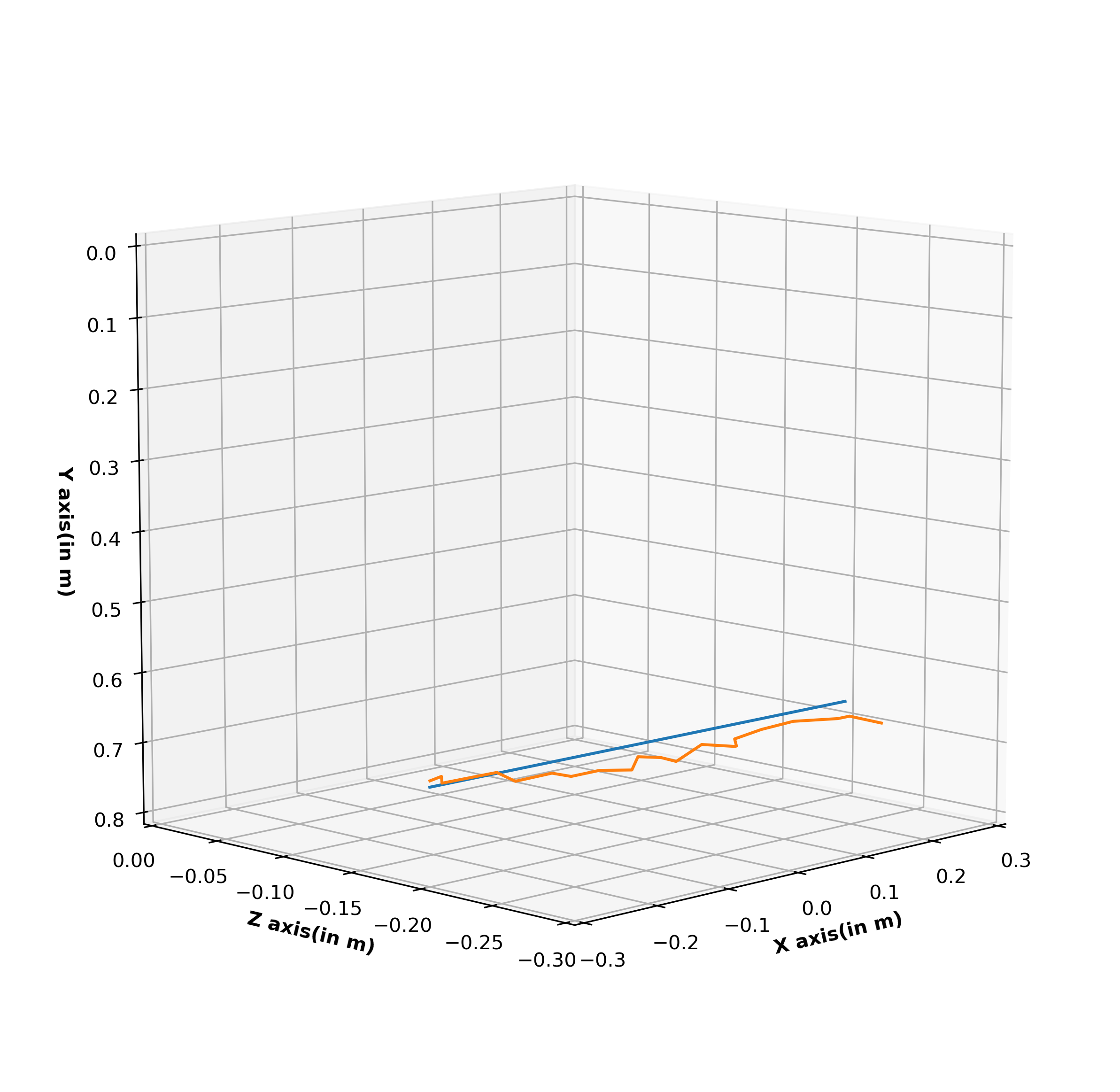}}
	\\
	\subfloat[]{\label{straightline0shot:2}\includegraphics[width=\columnwidth]{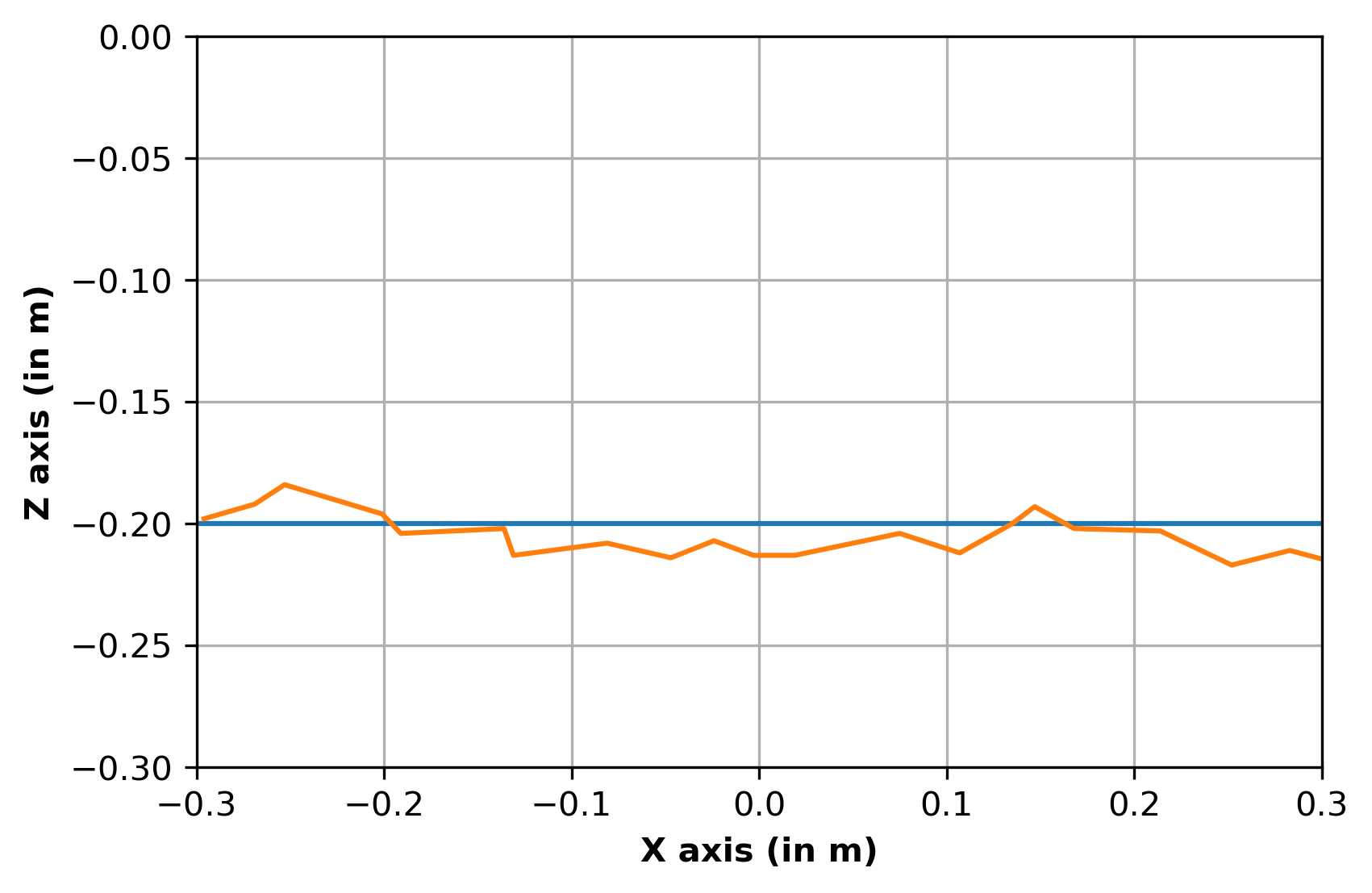}}
	\\
	\subfloat[]{\label{straightline0shot:2}\includegraphics[width=\columnwidth]{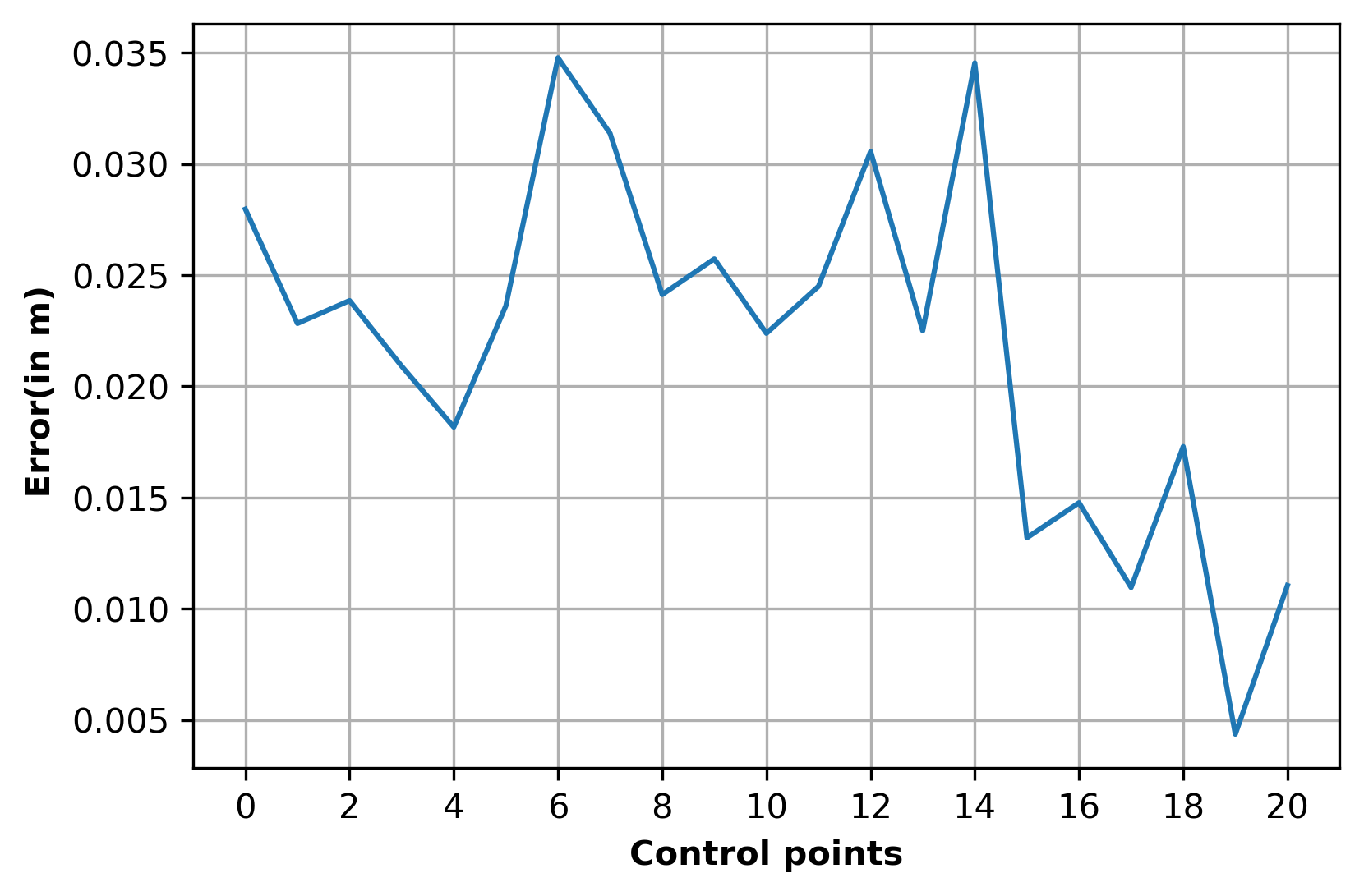}}
   \caption{Graph of trajectory 1 with known loading (a) 3D view (b) View in ZX plane (c) Tip positioning error. Blue and orange legends show desired trajectory and achieved trajectory respectively}
   \label{straightlineknown}
\end{figure}

\begin{figure}[htp]
	\centering
	\subfloat[]{\label{straightline0shot:1}\includegraphics[width=\columnwidth]{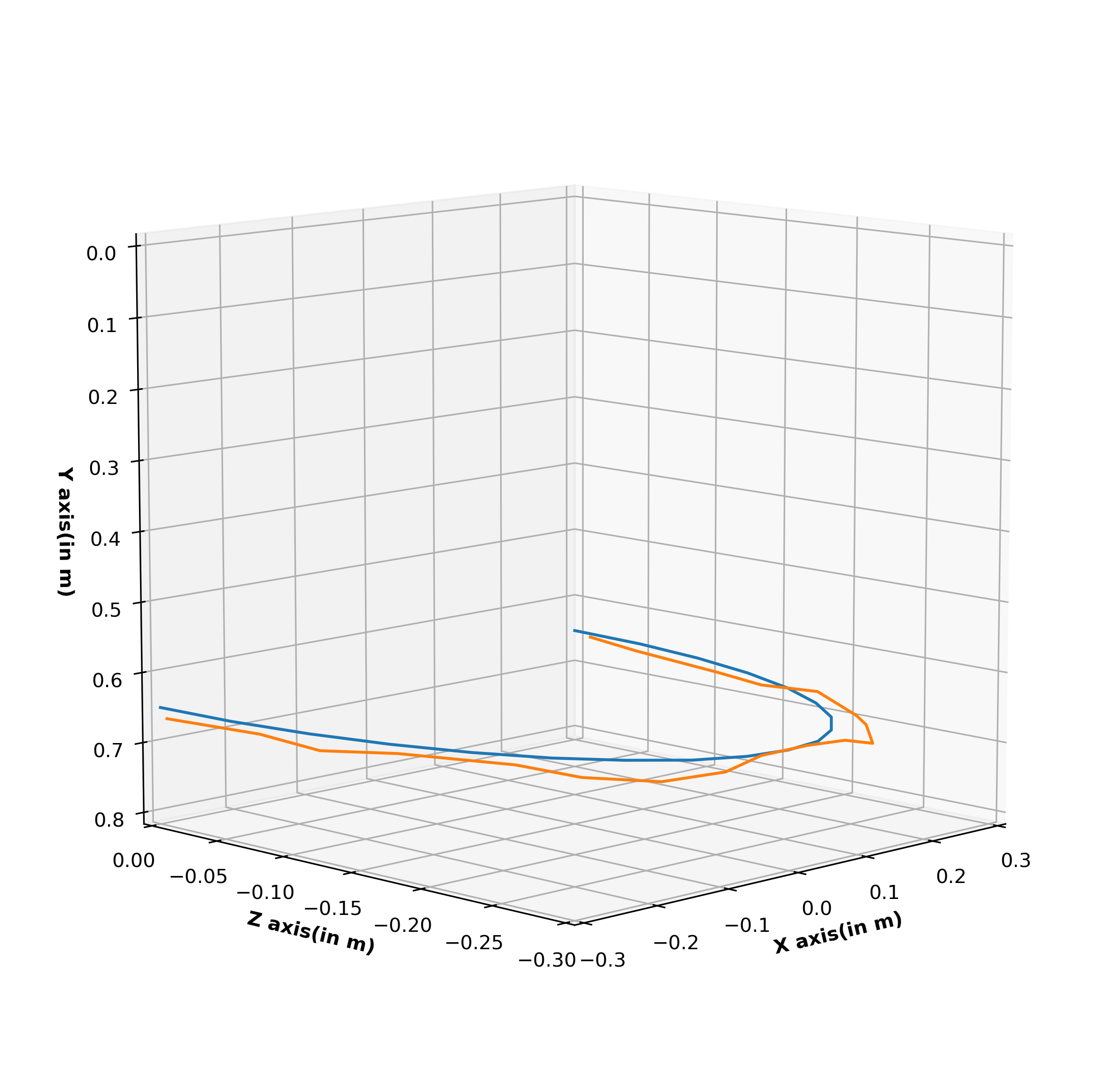}}
	\\
	\subfloat[]{\label{straightline0shot:2}\includegraphics[width=\columnwidth]{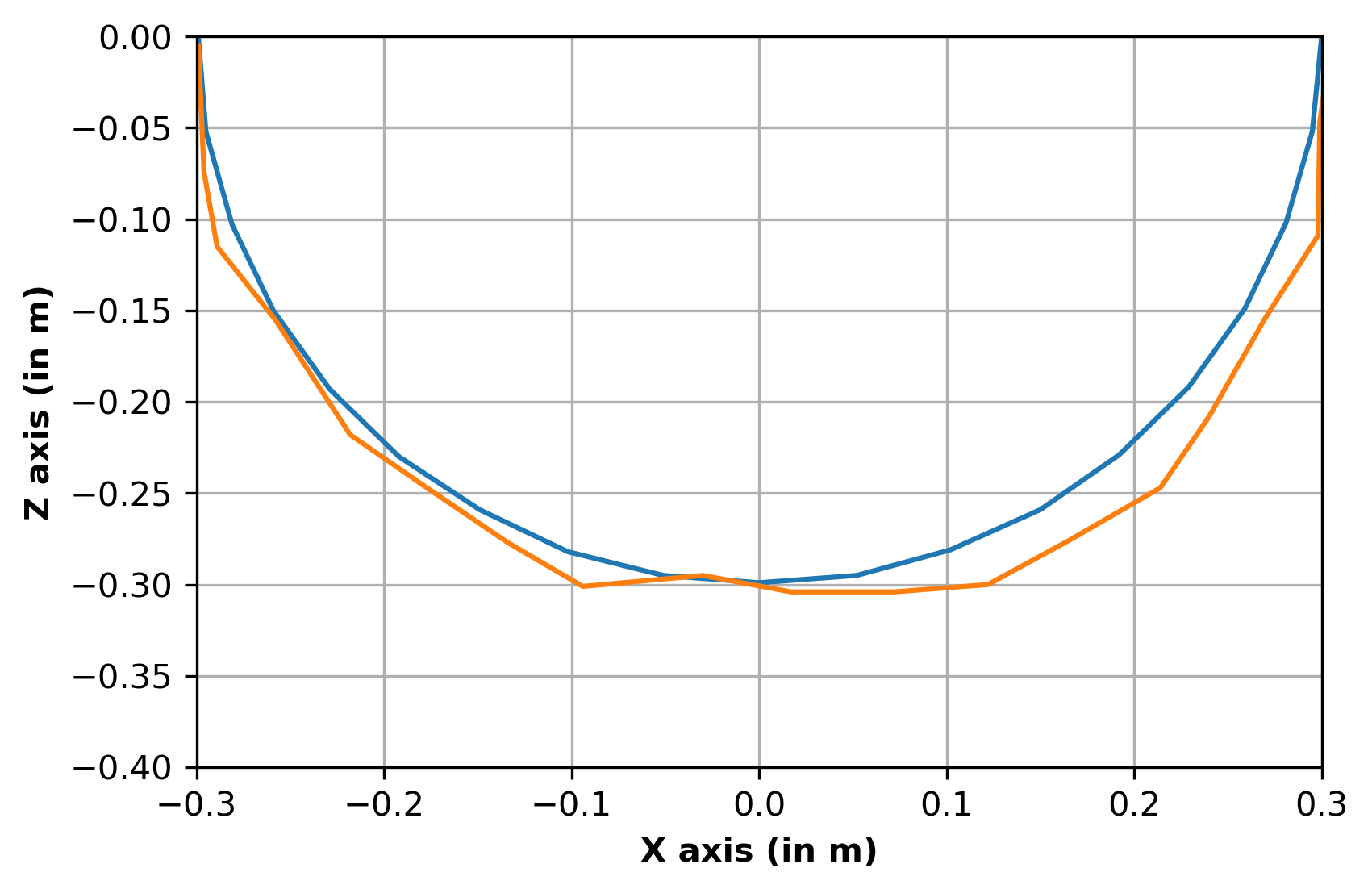}}
	\\
	\subfloat[]{\label{straightline0shot:2}\includegraphics[width=\columnwidth]{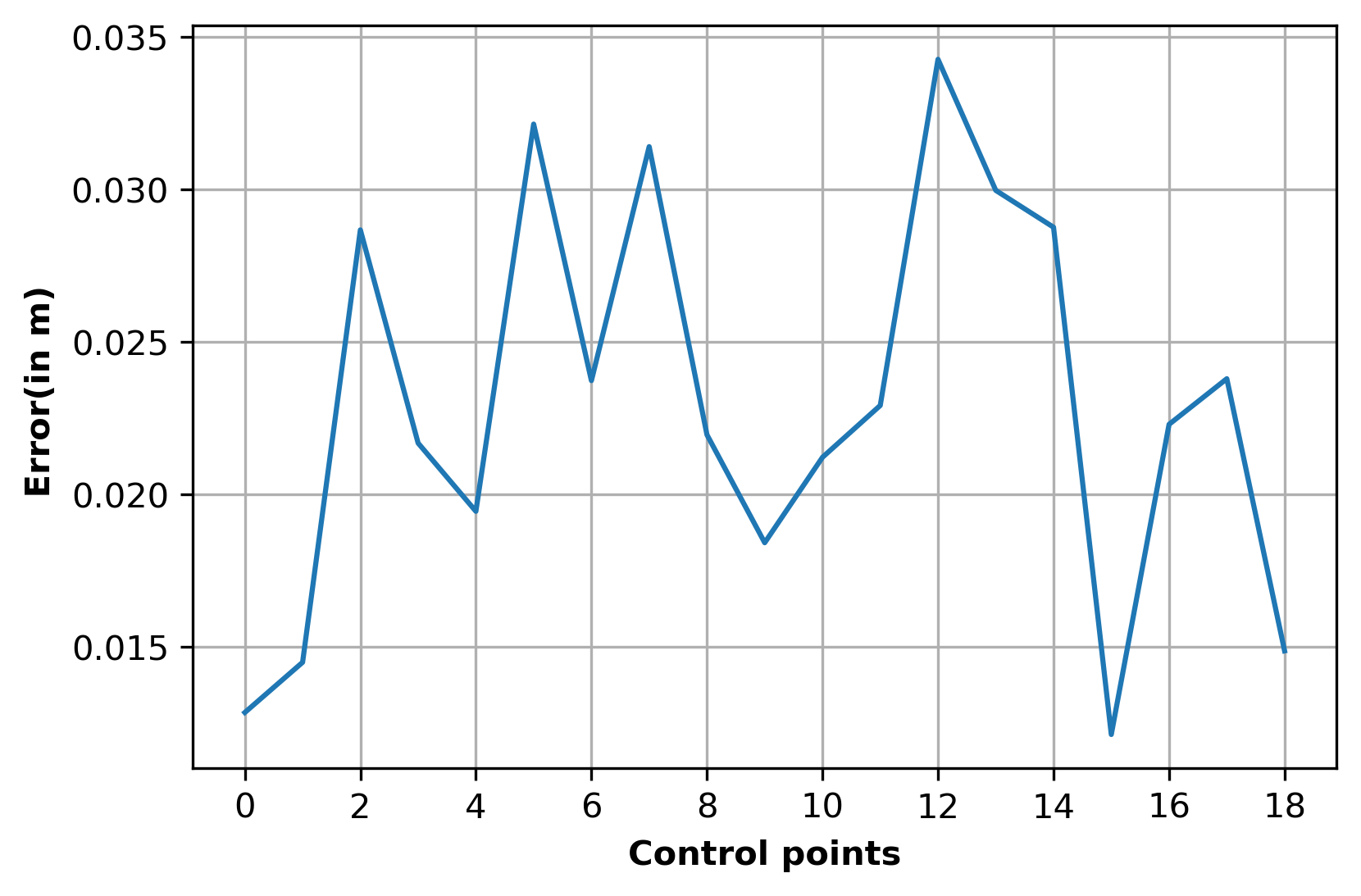}}
   \caption{Graph of trajectory 2 with known loading (a) 3D view (b) View in ZX plane (c) Tip positioning error. Blue and orange legends show desired trajectory and achieved trajectory respectively}
   \label{circleknown}
\end{figure}

\begin{table}
\caption{Results of trajectory following tasks with known external loading }
\label{trajectoryfollowing}
\setlength{\tabcolsep}{3pt}
\begin{tabular}{|p{70pt}|p{40pt}|p{40pt}|p{40pt}|p{40pt}|}
\hline
Task No. & Loading(in kg)& Tip positioning error(in m) & Standard deviation(in m) 
\\
\hline
Trajectory 1 & 0.2 & 0.0218 & 0.0097
\\
\hline
Trajectory 2 & 0.2 & 0.0227 & 0.0106
\\
\hline
\end{tabular}
\label{trajectory_following_performance_knownload}
\end{table}

From the figure \ref{straightlineknown}, we observe that the model took slightly zig-zag line. This phenomenon can be attributed to the fact that the distance between the control points are slightly larger. However, the average error for it remained below 0.0220 m. In the second trajectory, we observe the error was more than the straight line case. However, it was below 0.0230 m( relative error of below 3\%). So, we can see that our model is performing well for known loading case.  

\subsection{Trajectory following experiment with unknown external loading}

To test the capability of the model for unknown loading case, we used an external load of 0.250 kg. We used the same trajectory for this case as well. 
\begin{figure}[htp]
	\centering
	\subfloat[]{\label{straightline0shot:1}\includegraphics[width=\columnwidth]{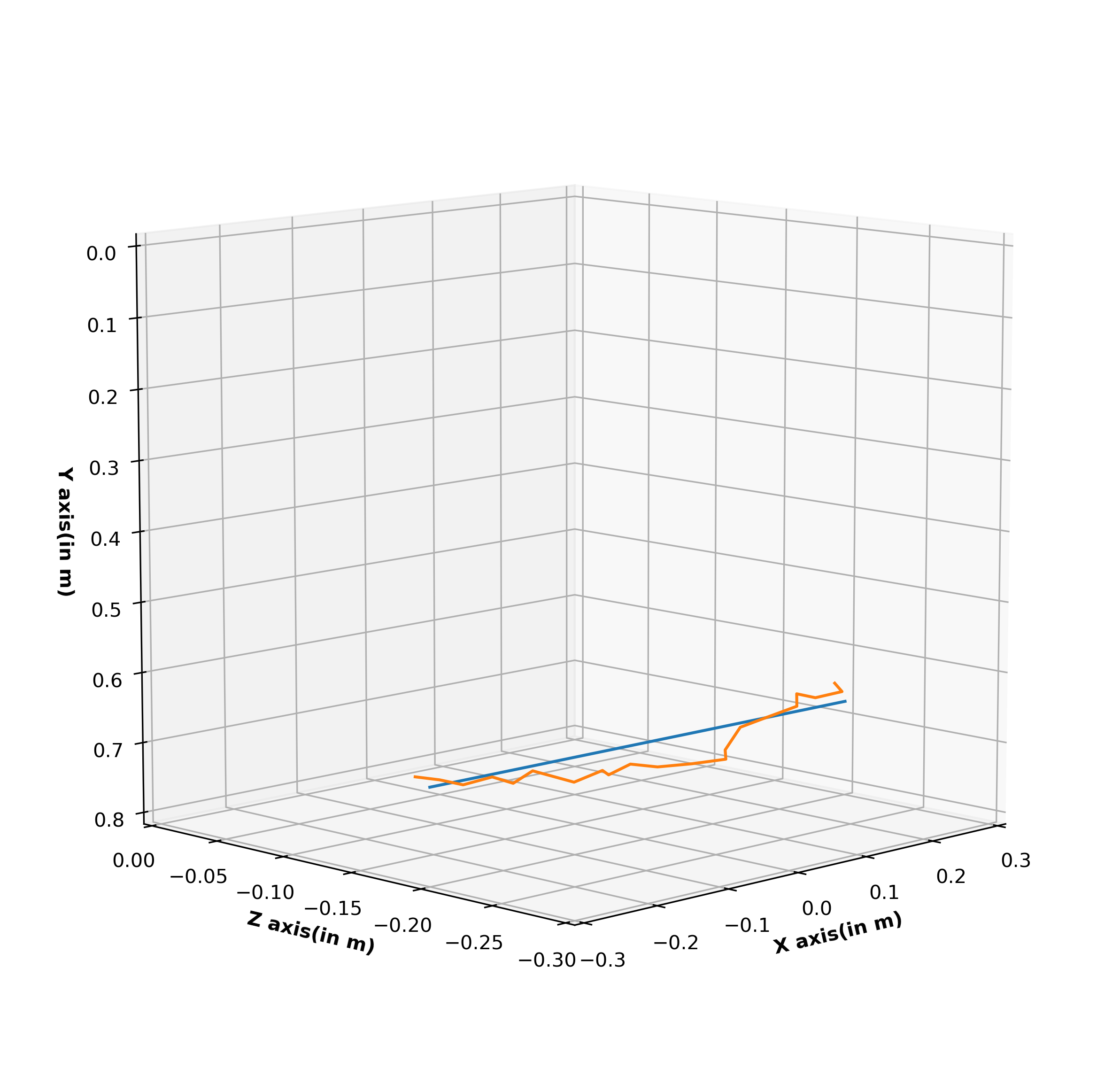}}
	\\
	\subfloat[]{\label{straightline0shot:2}\includegraphics[width=\columnwidth]{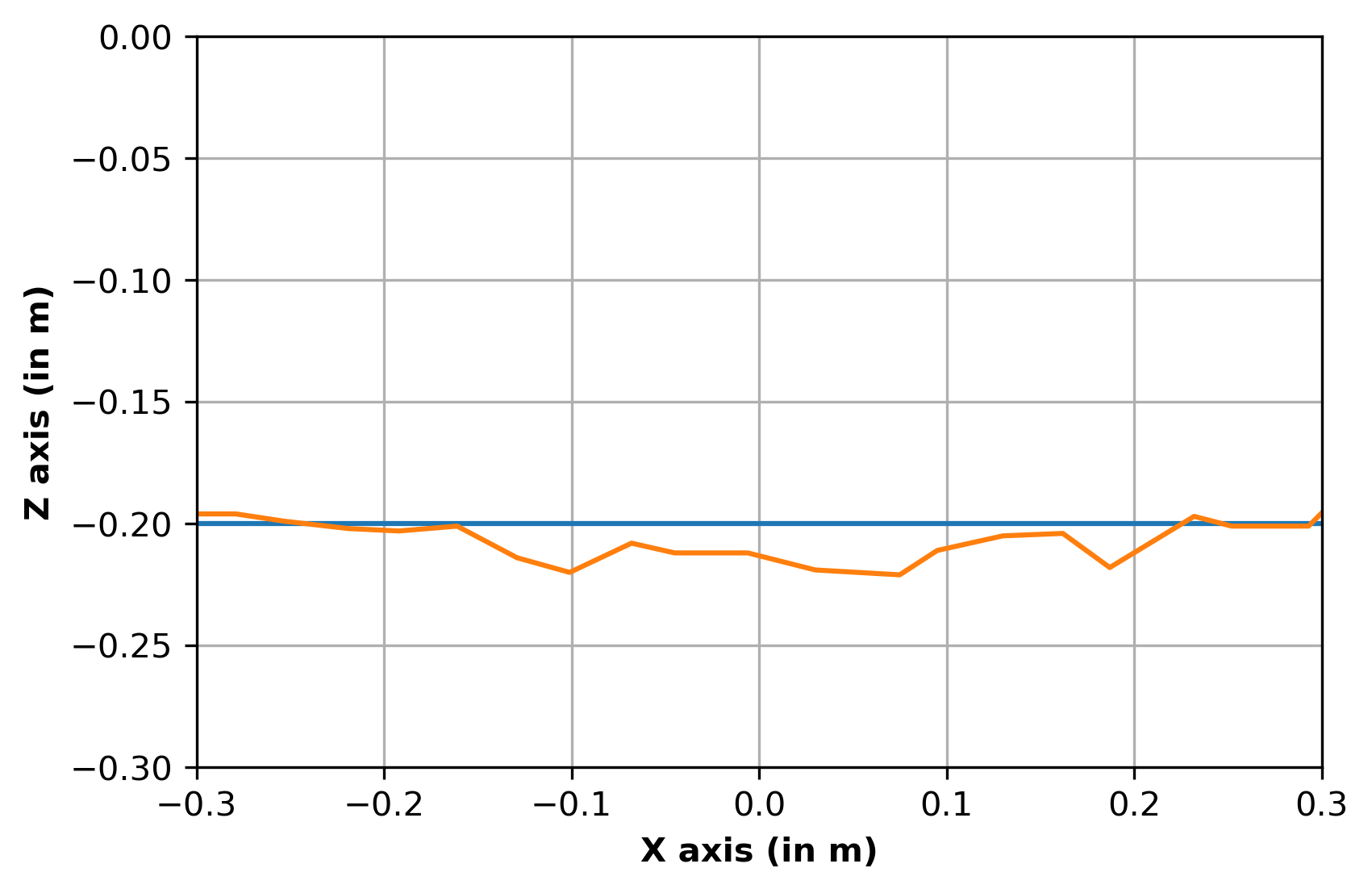}}
	\\
	\subfloat[]{\label{straightline0shot:2}\includegraphics[width=\columnwidth]{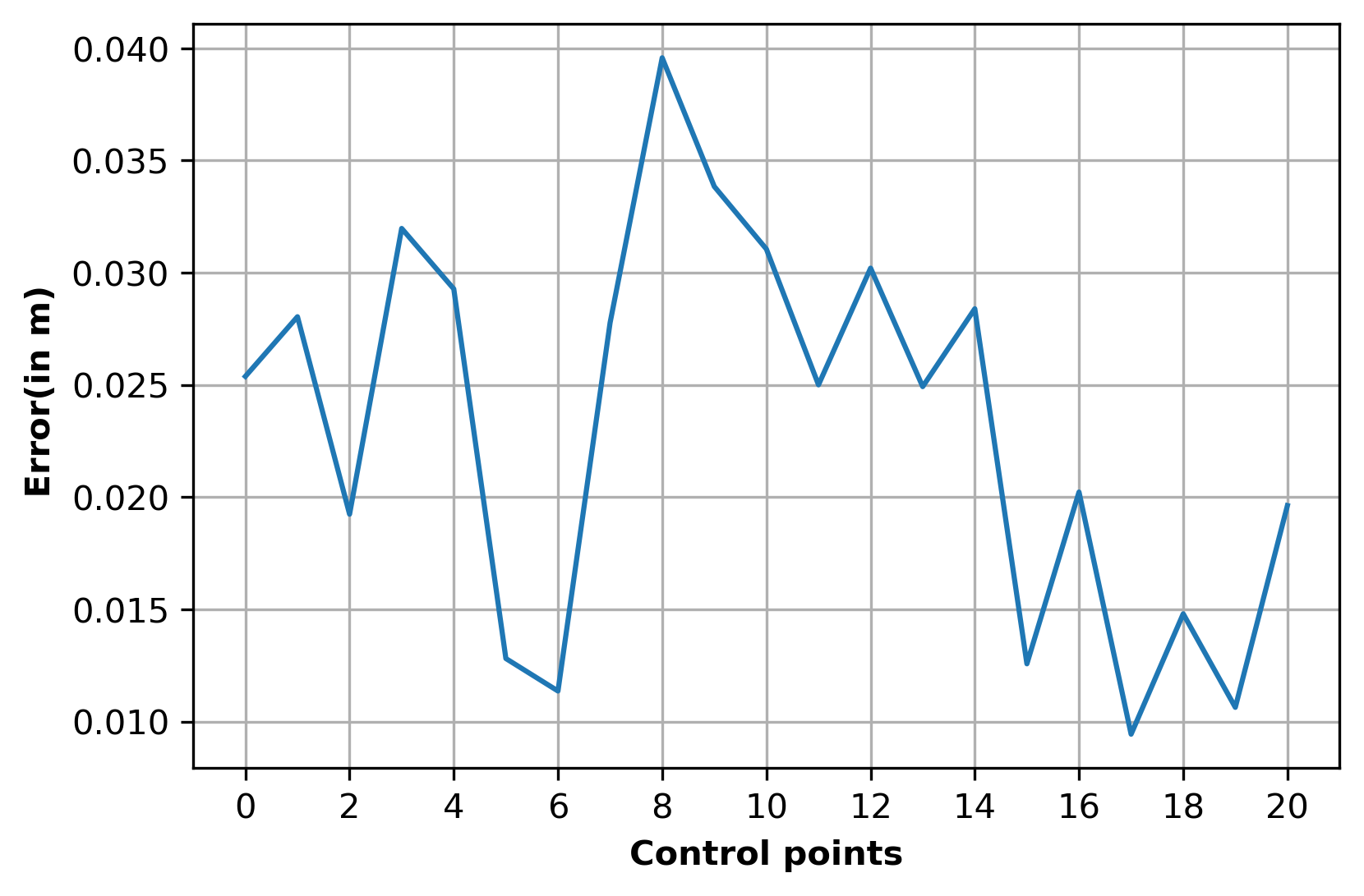}}
   \caption{Graph of trajectory 1 with unknown loading  (a) 3D view (b) View in ZX plane (c) Tip positioning error. Blue and orange legends show desired trajectory and achieved trajectory respectively}
   \label{straightlineunknown}
\end{figure}

\begin{figure}[htp]
	\centering
	\subfloat[]{\label{straightline0shot:1}\includegraphics[width=\columnwidth]{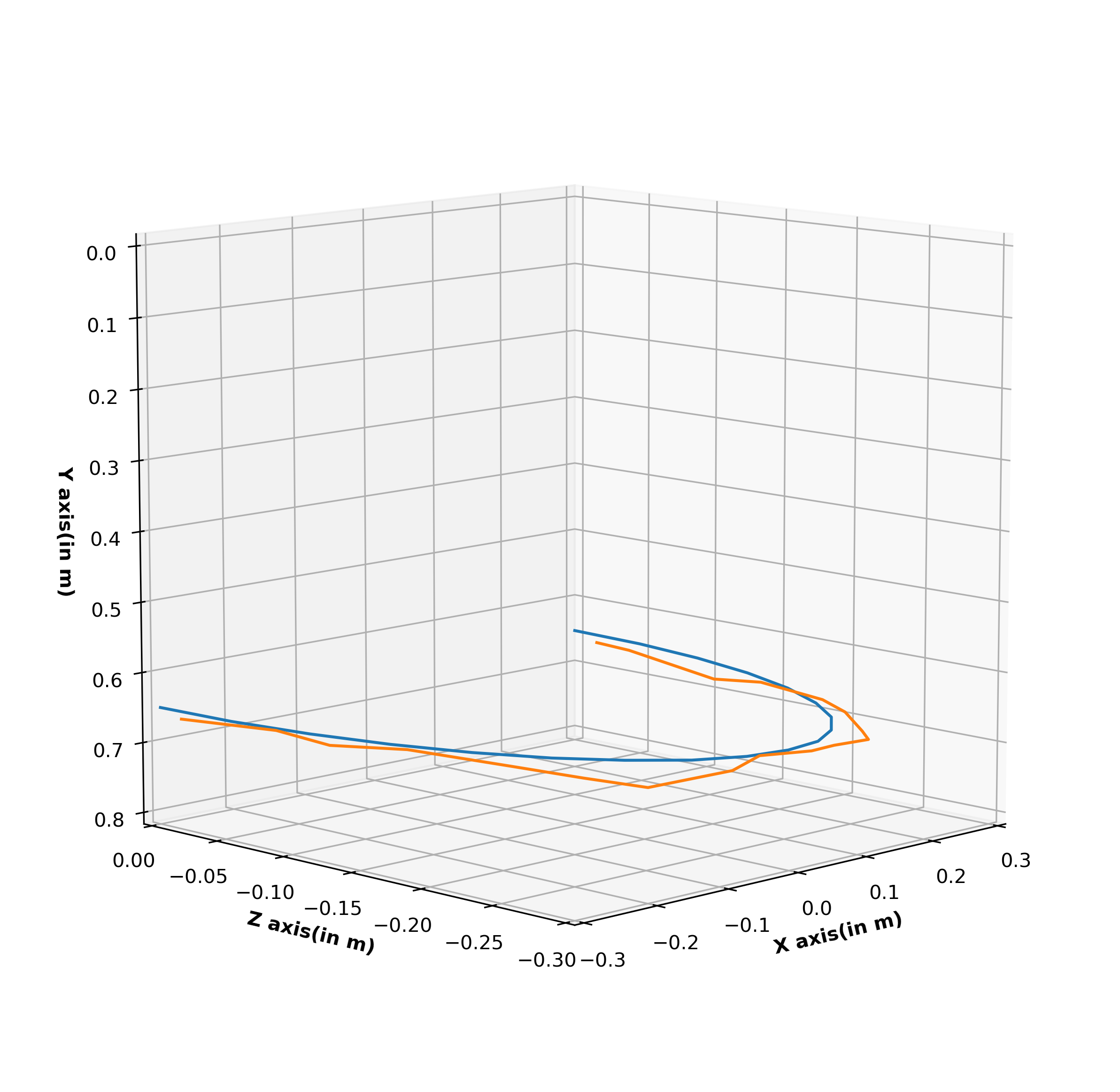}}
	\\
	\subfloat[]{\label{straightline0shot:2}\includegraphics[width=\columnwidth]{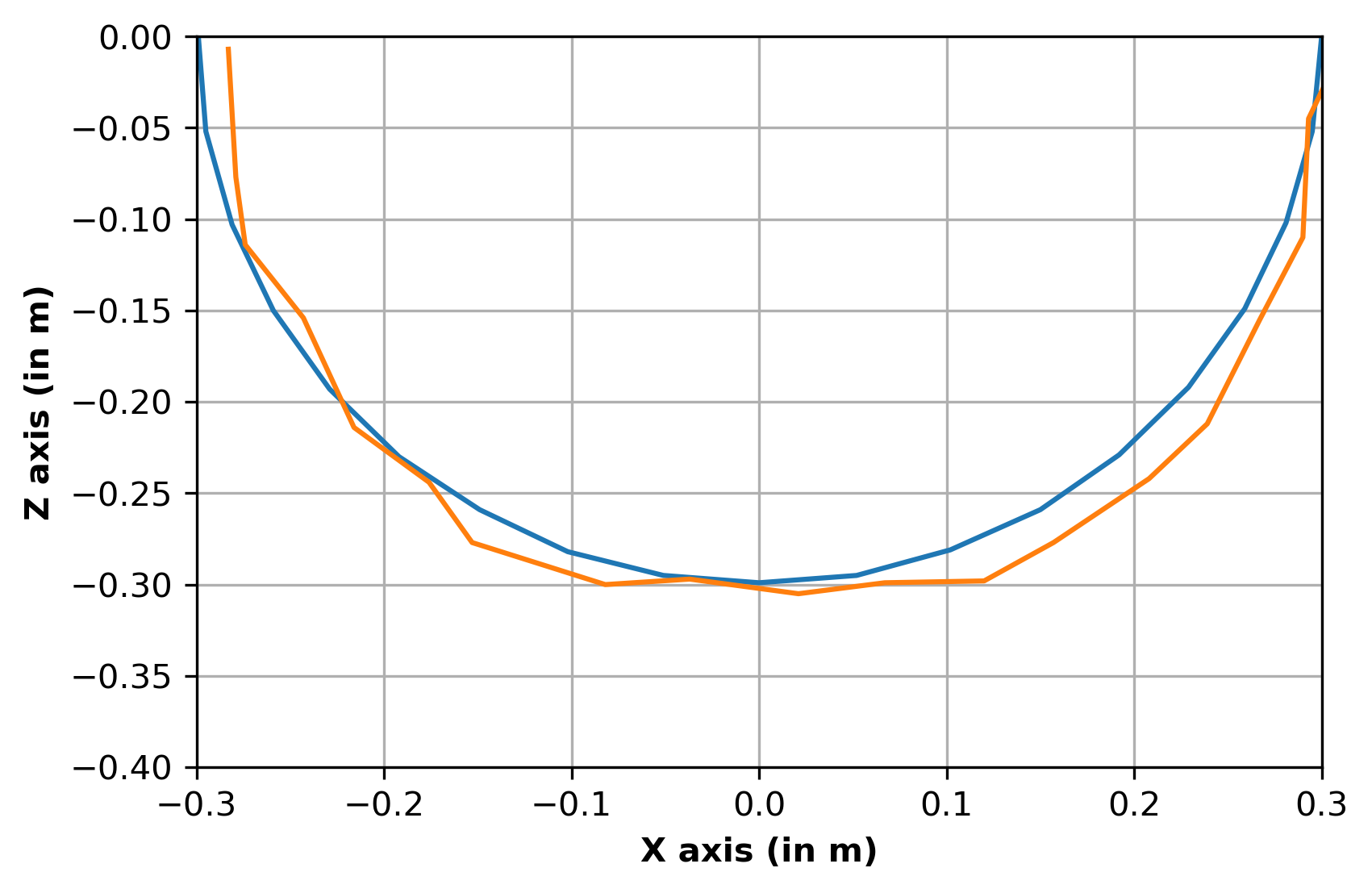}}
	\\
	\subfloat[]{\label{straightline0shot:2}\includegraphics[width=\columnwidth]{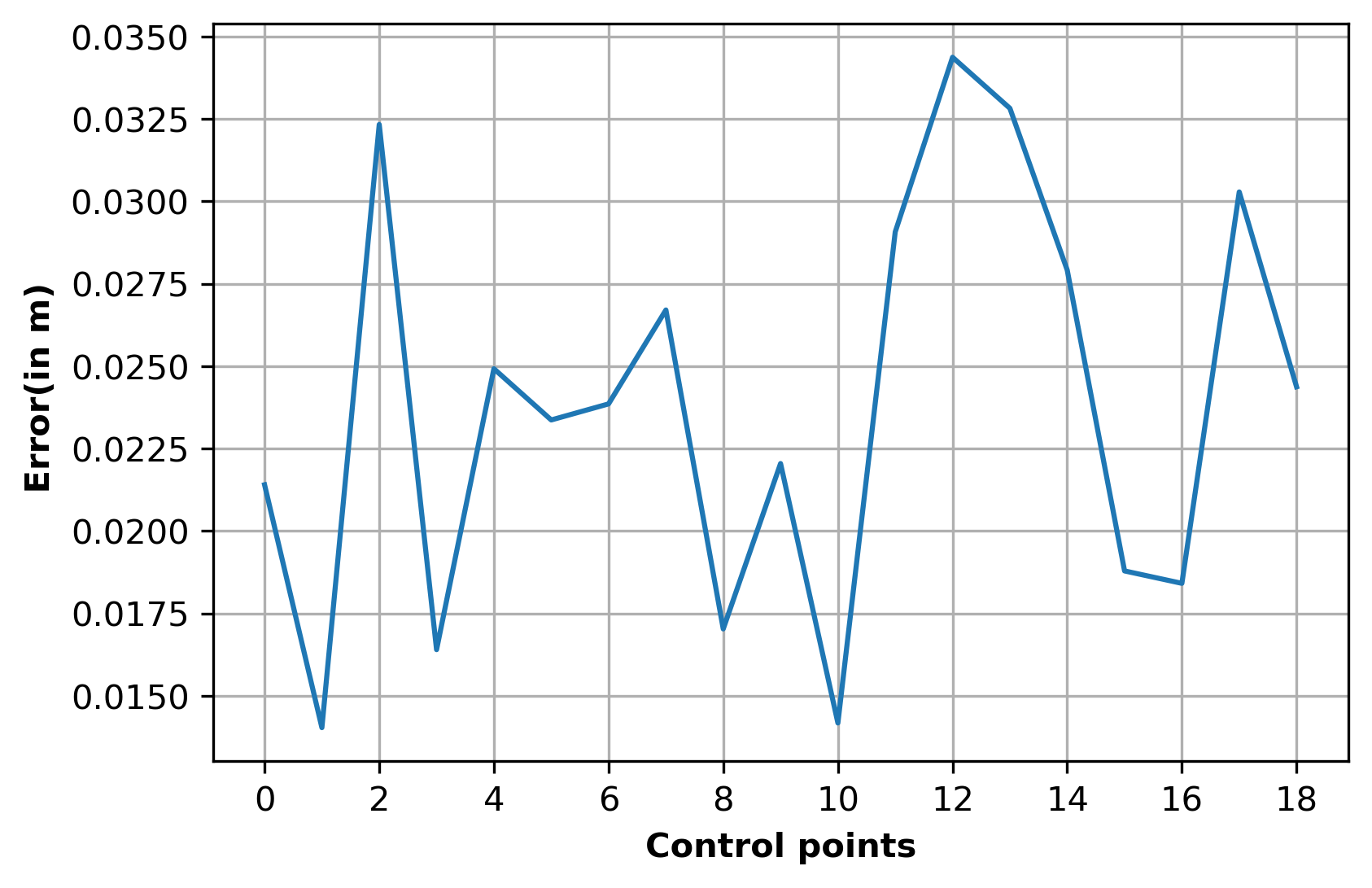}}
   \caption{Graph of trajectory 2 with unknown loading (a) 3D view (b) View in ZX plane (c) Tip positioning error. Blue and orange legends show desired trajectory and achieved trajectory respectively}
   \label{circleunknown}
\end{figure}

\begin{table}
\caption{  Trajectory following experiment with unknown external loading }
\label{trajectoryfollowing}
\setlength{\tabcolsep}{3pt}
\begin{tabular}{|p{70pt}|p{40pt}|p{40pt}|p{40pt}|p{40pt}|}
\hline
Task No. & Loading(in kg)& Tip error(in m) & Standard deviation(in m) 
\\
\hline
Trajectory 1 & 0.250 & 0.0231 & 0.0102
\\
\hline
Trajectory 2 & 0.250 & 0.0237 & 0.0098
\\
\hline
\end{tabular}
\label{trajectory_following_performance_unknownload}
\end{table}

From the figure \ref{straightlineunknown} and \ref{circleunknown}, it can be observed that we see a similar pattern as in case of known loading case. Here, the error was higher than the known case. However, the relative error is still below 3\%. 

\section{A comparative study}

We compared our result with the BPNN model proposed by Thuruthel et. el\cite{george2017learning} as we did in the last work. We trained the model with 10000 training data as before. Then , we tested it for the unknown loading case.

\begin{figure}[htp]
	\centering
	\subfloat[]{\label{straightline0shot:1}\includegraphics[width=\columnwidth]{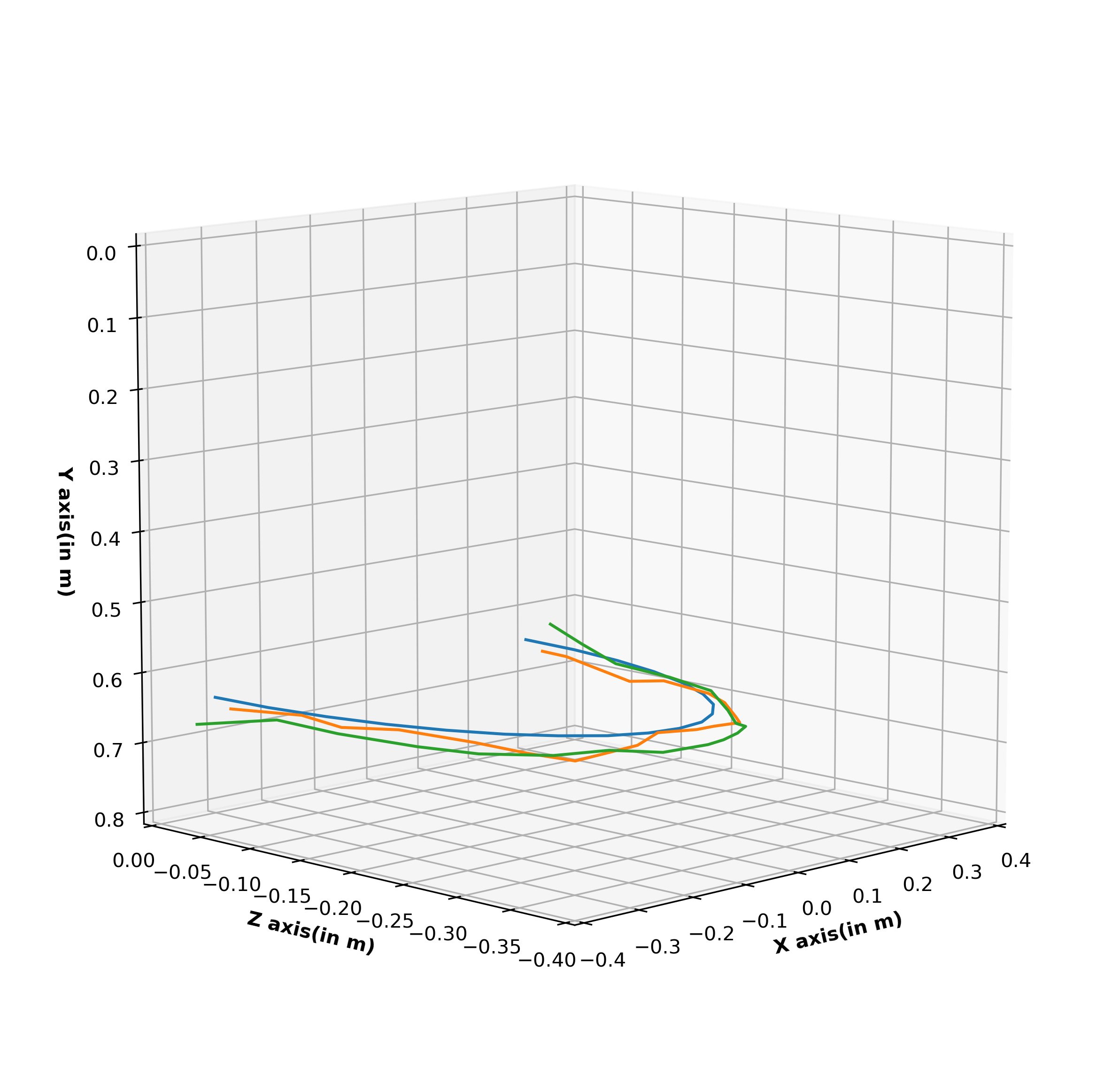}}
	\\
	\subfloat[]{\label{straightline0shot:2}\includegraphics[width=\columnwidth]{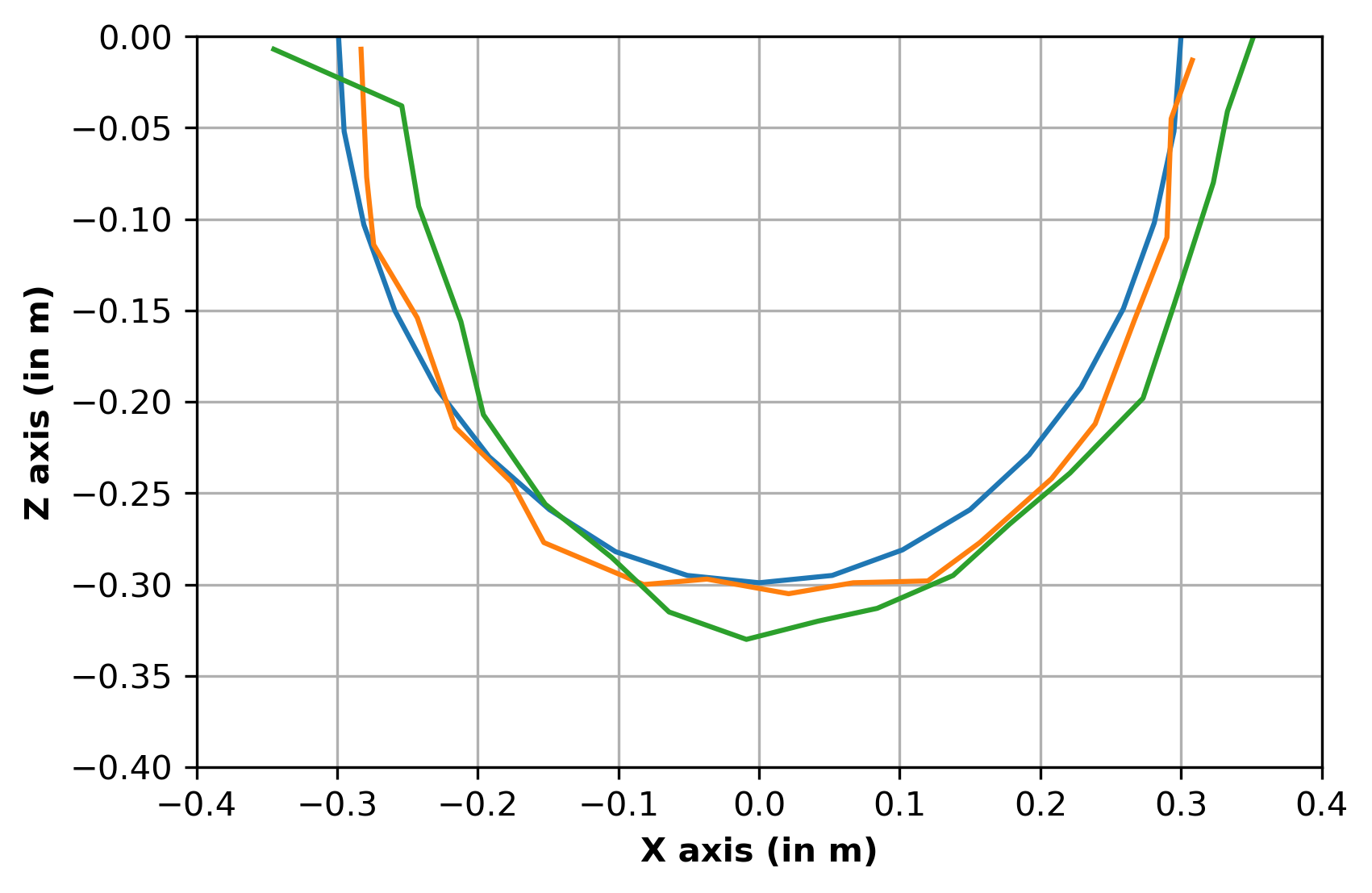}}
	\\
	\subfloat[]{\label{straightline0shot:2}\includegraphics[width=\columnwidth]{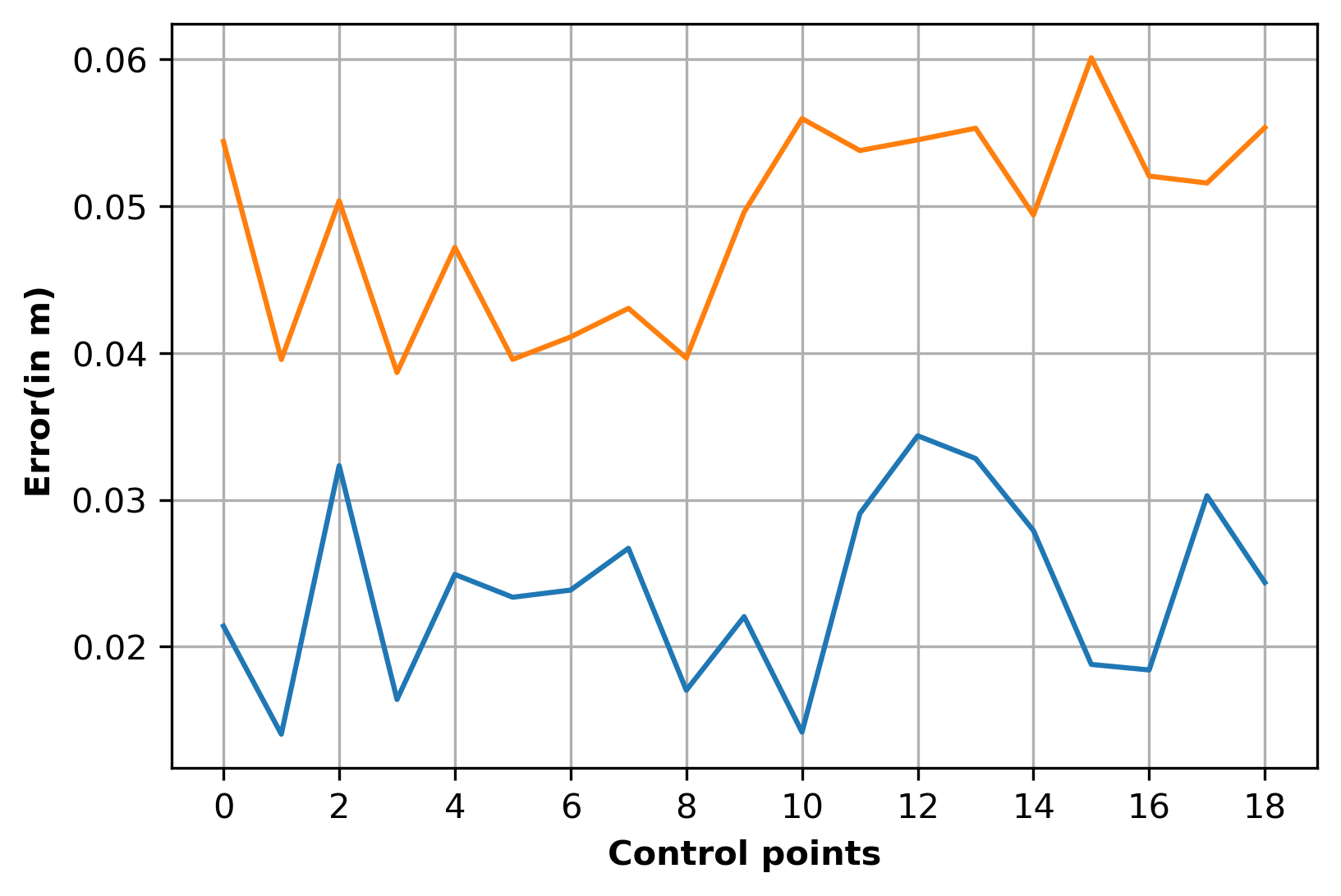}}
   \caption{Graph of trajectory 2 with unknown loading  (a) 3D view (b) View in ZX plane (c) Tip positioning error. Blue, orange and green legends show desired trajectory, achieved trajectory and achieved trajectory by our controller and BPNN respectively. }
   \label{circleunknowncomparision}
\end{figure}

\begin{table}
\caption{ Comparative results for the trajectory 2 with unknown external load}
\label{trajectoryfollowing}
\setlength{\tabcolsep}{3pt}
\begin{tabular}{|p{80pt}|p{50pt}|p{50pt}|p{50pt}|}
\hline
Algorithm & Loading(in kg)  & Tip error(in m) & Standard deviation(in m) 
\\
\hline
BPNN & 0.25  & 0.0491 & 0.0161
\\
\hline
MAML based adaptation & 0.25  & 0.0237 & 0.0098
\\
\hline
MAML based adaptation from CGAN data & 0.25  & 0.0256 & 0.0102
\\
\hline
\end{tabular}
\label{trajectory_following_performance_unknownload}
\end{table}

From the figure \ref{circleunknowncomparision}, it is evident that the model is performing poorly in the real environment. The error shoots above 0.06 m. The average error almost touches 0.05m. It could not adapt to the unknown load unlike last time where it tried to adapt to the new loading conditions. Now, it can be safely said that our model out performs the BPNN model by a huge margin. 

\section{Using GAN for data generation} 

From the above experiments, we saw that the simulation data is quite helpful for the initial training of the model. If we don't have a simulation model, then getting real data becomes the only option. GAN has currently shown a lot of success in creating fake datas in computer vision. Hence, we propose that a GAN can be used to create similar data sets using small number of real data sets. We are aware of the fact that GAN requires a huge amount of data to give good accuracy. However, we will try to train the model using small number of data as getting higher accuracy is not the main target here. The idea here is that we will use it for initial training only. We will use the same adaptation technique as discussed earlier.     
\subsection{A brief review of GAN}

Generative Adversarial Networks(GAN)s have shown great success in computer vision implementations. However, limited implementations have been done in continuum manipulator control. The basic idea of GAN is to let two neural networks (generator and discriminator) compete against each other. The work of the generator is to create fake data  similar to the original data. The job of the discriminator is to differentiate between real and fake data. 

Let us consider the original data distribution to be $p_o$. The generator should be able to find a data distribution $p_g$ close to the $p_o$. It is generally hard to get $p_r$ directly. Hence, this approach helps to get a proper data distribution\cite{ganreview}. 

Let us consider the discriminator and the generator have  parameters $\Phi_d$ and $\Phi_g$. First, the discriminator $Disc(x;\Phi_d)$ is trained using labeled real data and fake data. After it is able to differentiate, the generator,$Gen(x;\Phi_d)$, takes input from the latent space z. It tries to map it to a sample in the distribution of $p_g$. Then, the discriminator tries to differentiate between the real one and fake one. Based on the results from the discriminator, the generator tries to create more realistic fake data by giving a probability value $p(x;\Phi_d)$ . The over all objective function for it is \cite{goodfellow2016deep}


\begin{equation}
 \begin{aligned}
  min_{Gen} max_{Disc} V(Disc,Gen)=
  E_{X\sim p_{data}(x)}[log p(x)] + \\E_{x\sim p_{model}}[log(1-p(x))]
  \end{aligned}
\end{equation}

Here, V(Disc,Gen) determines the payoff of the discriminator. Generator receives -V(Disc,Gen) as its own payoff. 
This converges when the discriminator is no longer capable to differentiate between real and fake data. It outputs 0.5 then. After convergence, generator is used to create fake data.

However, it is seen that GANs are quite unstable and it may suffer from vanishing gradient problem. It also collapses where the generator outputs nearly similar output.
\subsection{CGAN}
The main idea of conditional GAN is that a support information vector y is provided here. It is available to both discriminator and generator. It helps to converge faster. Moreover, we can have the required output by the generator by using the information. In our case, the information vector is the external load. Here, we first gave 13 random actuator values for each external loading in the range of [0,0.5] with a difference of 0.1 kg. The actuator values and corresponding tip point values were recorded for those trajectories. 30 values in regular interval between the starting input-output pairs were treated as the data points for training(excluding starting point). Given those points, we used a simple technique to increase the data points. We created random values between [-50000,50000]. We added those randomly to the existing encoder values. We also changed the tip position by randomly adding values between [-0.01,0.01] randomly in x, y and z directions. We increased the number of data points 6 times using this technique. 
\subsection{Error in the CGAN data}

\begin{figure}[!t]
\centerline{\includegraphics[width=\columnwidth]{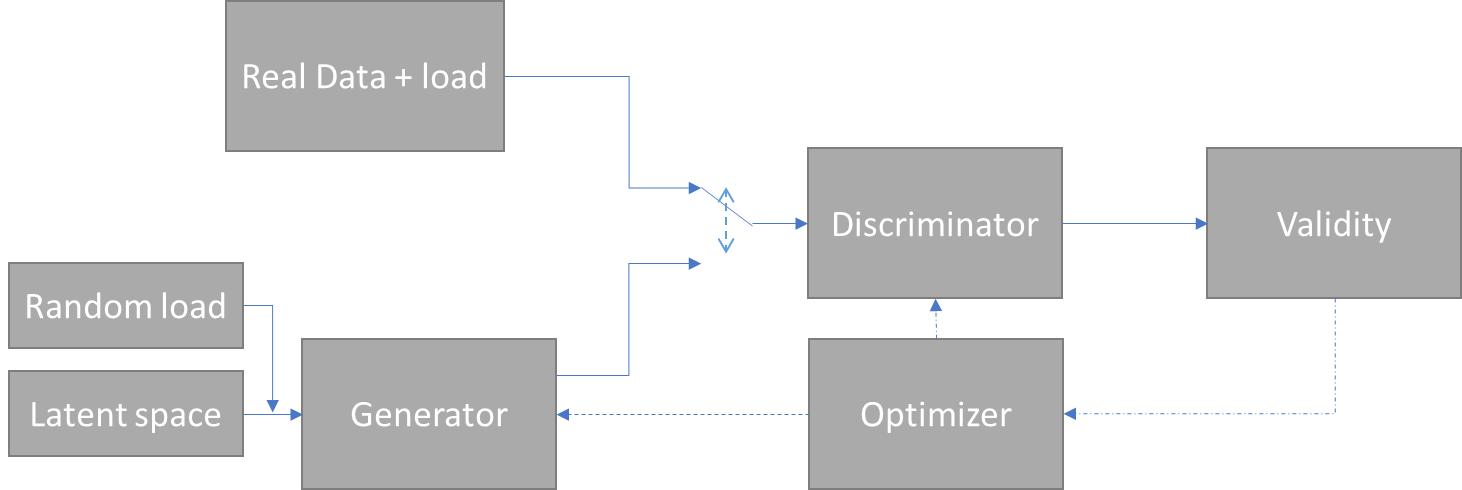}}
\caption{Proposed Conditional GAN}
\label{cgan}
\end{figure}

\begin{table}
\caption{Details of GAN network training}
\label{table1}
\setlength{\tabcolsep}{3pt}
\begin{tabular}{|p{100pt}|p{100pt}|}
\hline
No. of layers & 2
 \\
\hline
Hidden layer size & 256
\\
\hline
Activation function & ReLU
\\
\hline
Epochs & 200
\\
\hline
loss & binary cross-entropy 
\\
\hline
optimizer & adam 
\\
\hline
Latent space dimension & 50
\\
\hline
\end{tabular}
\label{traingperformance}
\end{table}

To test the accuracy of the generated data points, we took 50 data points having (($p_{next}$,$a_{curr}$,$p_{curr}$) $\rightarrow$ $a_{next}$) values. To verify its accuracy,we did a random point test. We took 50 random data points generated by the generator. Then, we tested the tip positioning error for it on the real robot. We found the average error to be 0.0351$\pm$ 0.0115m.  We generated 8000 data points from the generator. Now, we used these data set with the original data set to train the neural network.  
\begin{table}[h]
\caption{Results of the the model trained with CGAN data }
\label{table}
\setlength{\tabcolsep}{3pt}
\begin{tabular}{|p{50pt}|p{50pt}|p{50pt}|p{50pt}|}
\hline
Number of gradient steps &
Average error(in m) & standard deviation(in m) 
 \\
\hline
0& 0.0336 & 0.0198
\\
\hline
1& 0.0289 & 0.0172
\\
\hline
2& 0.0247 & 0.0185
\\
\hline
3& 0.0235 & 0.0135
\\
\hline
4& 0.0232 & 0.0129
\\
\hline
\end{tabular}
\label{cganbasedresult}
\end{table}
\subsection{Experiment with CGAN-MAML model}
We used the controller directly to the real environment first. It was found out that the average error for it was 0.0336$\pm$ 0.0193 at first. It is comparatively less than the result we got from the model trained with simulation data.  However, from the table \ref{cganbasedresult}, we find that the current model is slower in adapting to the actual world in comparison to the previous model trained with simulated data. It may be due to the fact that the previous model was trained with the data having higher distribution. The current model was trained using more similar data. Hence, it is always better to have a simulation model. However,if the simulation model is not available and collecting real time data is expensive, then we can use the above proposed method.

\subsection{A comparative study}

\begin{figure}[htp]
	\centering
	\subfloat[]{\label{straightline0shot:1}\includegraphics[width=\columnwidth]{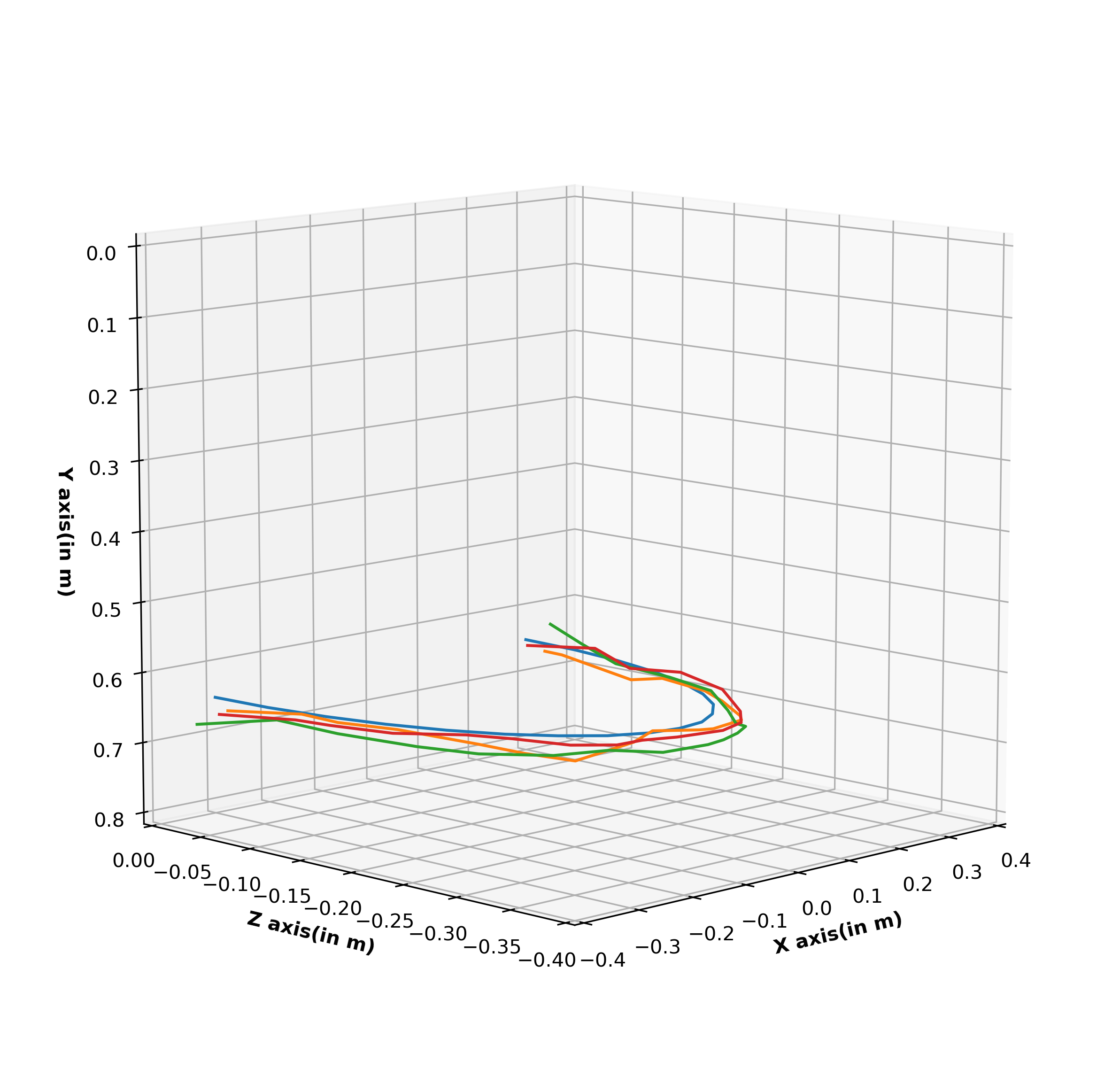}}
	\\
	\subfloat[]{\label{straightline0shot:2}\includegraphics[width=\columnwidth]{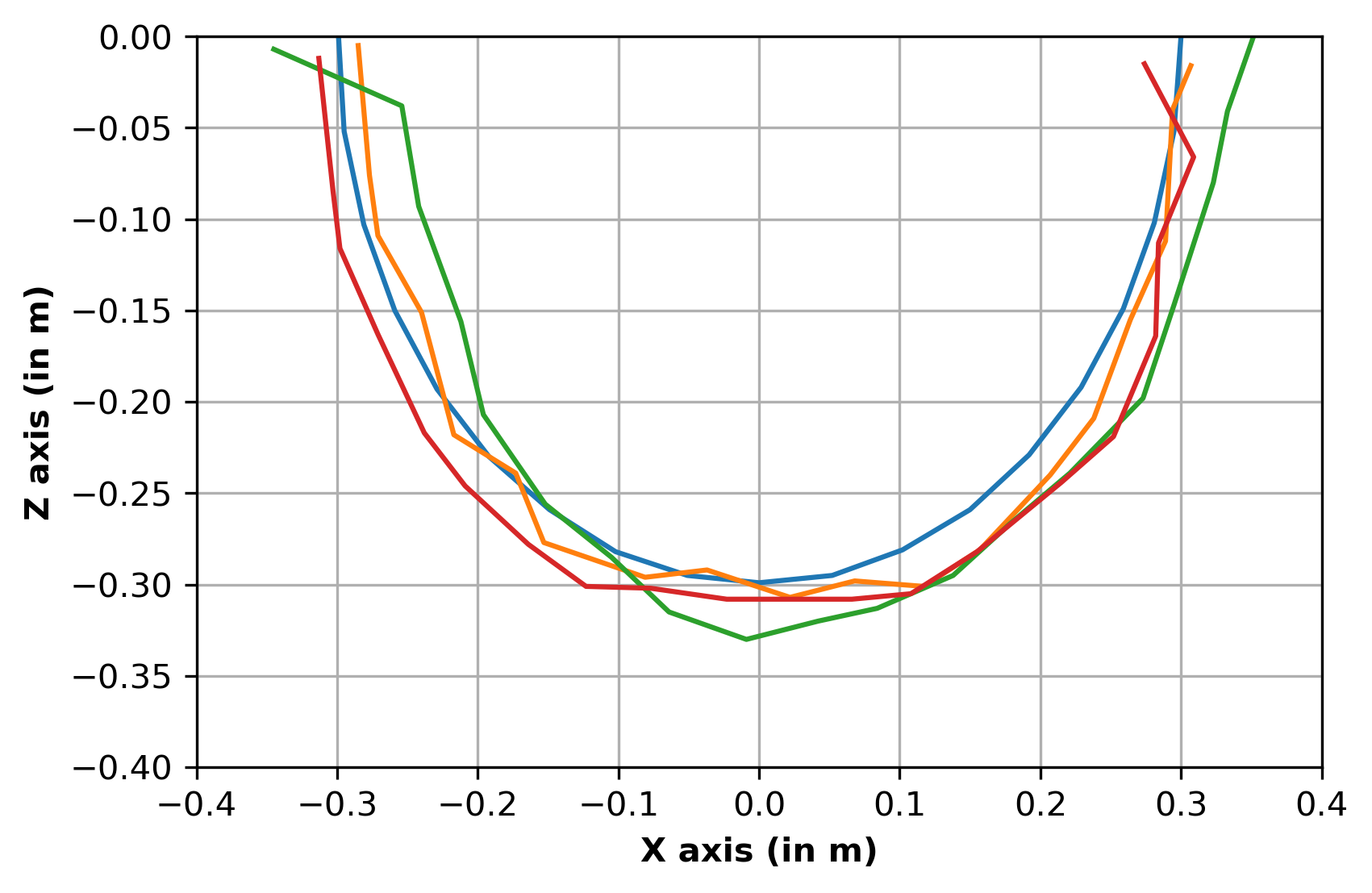}}
	\\
	\subfloat[]{\label{straightline0shot:2}\includegraphics[width=\columnwidth]{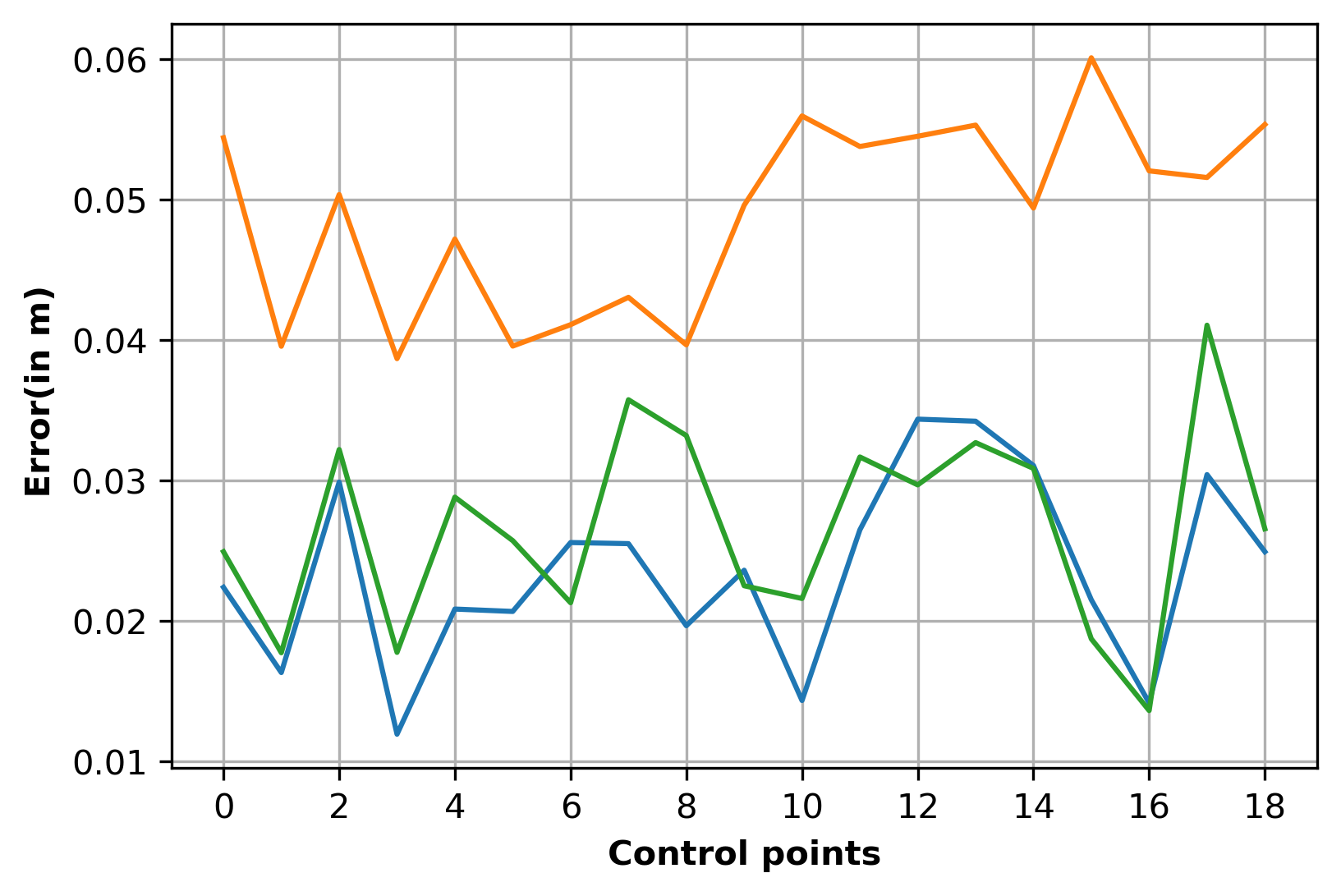}}
  \caption{Graph of trajectory 2 with unknown loading (a) 3D view (b) View in ZX plane (c) Tip positioning error . The blue, orange, green and red legends show the desired trajectory, achieved trajectory using our controller, BPNN and cgan data trained controller respectively.}
  \label{circleunknowncomparisioncgan}
\end{figure}
We further tried to see its performance with respect to previous controllers. Hence, we choose to see its performance for trajectory following task. We choose trajectory-2 for it. Here, we will consider unknown loading case. From the figure \ref{circleunknowncomparisioncgan}, we can observe that model trained with data from cgan performed better then BPNN model from the last experiment. However, it could not surpass the performance of the model trained with simulation data. As explained earlier, it must be due to the models training using similar data. However, the average error was below 0.03m (0.0256 m). Hence, it can be used for the cases where we do not have a simulation model. We can further improve the model's accuracy using more data points to train the CGAN. Further experiments can be done to find out exact ratio of real to synthetic data needed to achieve better accuracy.

\section{Discussions and Conclusions}

In this paper, we proposed a MAML based sim-to-real approach. First, we trained the controller using simulation model based on 3D mechanics model of the robot. Then, we used the adaptation step to adapt to the real world. It was observed that the model trained with only simulation data points required 5 gradient steps to adapt to the real environment. The relative errors were below 2.94\% for known cases and 3.07 \% for unknown loading cases. Thuruthel et. el\cite{george2017learning,thuruthel2018stable} were successful to achieve relative error up to 2.76 percent using model based reinforcement learning\cite{thuruthel2018model}. Deep reinforcement learning based controller of Satheeshbabu et. el \cite{satheeshbabu2020continuous} achieved a relative error of 3.87 percent. We further compared the model's performance with BPNN model. It was found that our model out performs the BPNN model. Then, we proposed a CGAN based method for the cases where simulation model is not available. We trained the CGAN model with the data from generated 1800 real data points. Then, we generated fake data for the training using CGAN. After adaptation step, the relative error for it was below 3\%.  Finally, we can come to the conclusion that we can use both simulator data and cgan generated low accuracy data for training and use MAML based adaptation in the real world.


\bibliographystyle{ieeetr}
\bibliography{references.bib}
\begin{IEEEbiography}[{\includegraphics[width=1in,height=1.25in,clip,keepaspectratio]{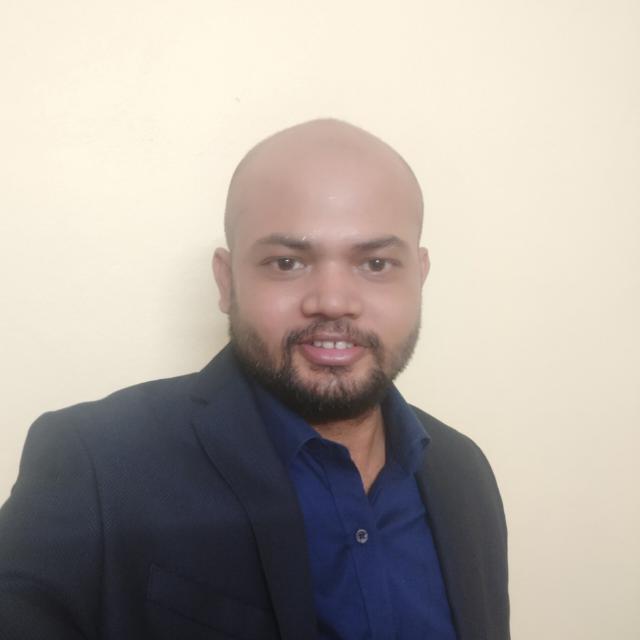}}]{ALOK RANJAN SAHOO }  received the B.Tech degree in Mechanical engineering. He received the M.Tech degree in Robotics from IIIT Allahabad. He is currently pursuing PhD in Robotics from IIIT Allahabad. His research interests are robotics, machine learning and computer vision.
\end{IEEEbiography}

\begin{IEEEbiography}[{\includegraphics[width=1in,height=1.25in,clip,keepaspectratio]{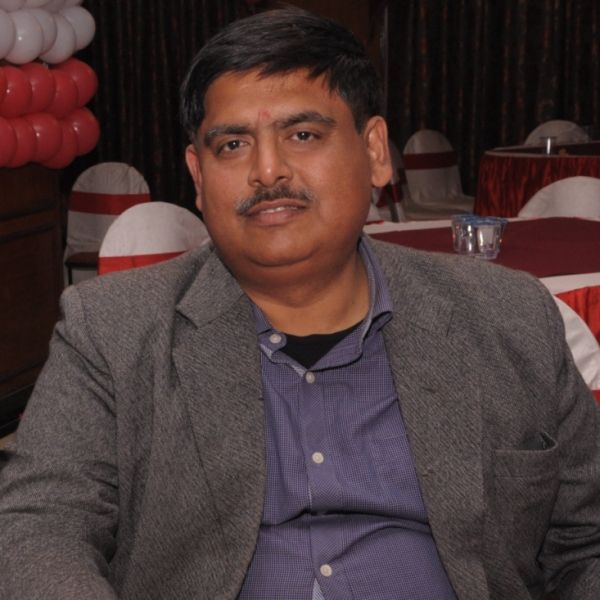}}]{PAVAN CHAKRABORTY } received the Ph.D.
degree from the Indian Institute of Astrophysics,
India, in 2001. He is currently a Professor
with the Indian Institute of Information
Technology Allahabad. His research interests
include human gait analysis, human prosthetic,
bio-metrics, image processing, graphics and visual
computing, graphical projections, robotics and
instrumentation, real time simulation, high performance
computing (HPC), artificial life simulation
and intelligence, and human-computer interaction.
\end{IEEEbiography}

\end{document}